\pdfoutput=1

\documentclass[11pt]{article}

\usepackage[]{EACL2023}

\usepackage{times}
\usepackage{latexsym}
\usepackage{graphicx}
\usepackage{booktabs}
\usepackage{array}
\usepackage{makecell}
\usepackage{multirow}
\usepackage{float}
\usepackage{tabularx}
\usepackage{enumitem}
\usepackage{comment}
\usepackage{fdsymbol}
\usepackage[justification=centering]{caption}
\usepackage[T1]{fontenc}

\usepackage[utf8]{inputenc}

\usepackage{microtype}

\usepackage{inconsolata}

\newcommand{\ignore}[1]{}

\title{Simple Yet Effective Synthetic Dataset Construction for \\ Unsupervised Opinion Summarization}

\author{Ming Shen\textsuperscript{{$\varheartsuit$}}\thanks{\hspace{5pt}Work done during an internship at AWS AI Labs.} \quad Jie Ma\textsuperscript{{$\clubsuit$}}\quad Shuai Wang\textsuperscript{{$\clubsuit$}}\quad Yogarshi Vyas\textsuperscript{{$\clubsuit$}}\quad \\ \textbf{Kalpit Dixit\textsuperscript{{$\clubsuit$}}\quad Miguel Ballesteros\textsuperscript{{$\clubsuit$}}\quad Yassine Benajiba\textsuperscript{{$\clubsuit$}}} \\
{\textsuperscript{$\varheartsuit$}Arizona State University} \quad {\textsuperscript{$\clubsuit$}AWS AI Labs} \\
{\texttt{mshen16@asu.edu;\{jieman,wshui,yogarshi,kddixit,ballemig,benajiy\}@amazon.com}}
}

\begin{document}
\maketitle

\begin{abstract}


Opinion summarization provides an important solution for summarizing opinions expressed among a large number of reviews. However, generating aspect-specific and general summaries is challenging due to the lack of annotated data. In this work, we propose two simple yet effective unsupervised approaches to generate both aspect-specific and general opinion summaries by training on synthetic datasets constructed with aspect-related review contents. 
Our first approach, \textit{Seed Words Based Leave-One-Out} (\textsc{SW-LOO}), identifies aspect-related portions of reviews simply by exact-matching aspect seed words and outperforms existing methods by $3.4$ ROUGE-L points on \textsc{Space} and $0.5$ ROUGE-1 point on \textsc{Oposum+} for aspect-specific opinion summarization.
Our second approach, \textit{Natural Language Inference Based Leave-One-Out} (\textsc{NLI-LOO}) identifies aspect-related sentences utilizing an NLI model in a more general setting without using seed words and outperforms existing approaches by $1.2$ ROUGE-L points on \textsc{Space} for aspect-specific opinion summarization and remains competitive on other metrics. 

\end{abstract}

\section{Introduction}
\label{sec:intro}
Customer reviews play a vital role in decision-making for customers and product (or business) providers, as customers usually resort to reviews to guide their purchasing decisions and product providers improve their products based on reviews as feedback. However, it becomes hard for customers or product providers to read through all reviews before making decisions with the explosion of online reviews in recent years. Opinion summarization \cite{hu2006opinion, wang-ling-2016-neural, angelidis-lapata-2018-summarizing, brazinskas-etal-2020-unsupervised,  brazinskas-etal-2022-efficient, angelidis-etal-2021-extractive, amplayo-etal-2021-aspect, basu-roy-chowdhury-etal-2022-unsupervised}, the task of generating a general summary of \textit{salient} opinions expressed among reviews, provides a feasible solution to this problem. 

Different from summarization in Wikipedia and news domains \cite{nallapati-etal-2016-abstractive, narayan-etal-2018-dont, see-etal-2017-get, narayan-etal-2018-ranking, liu-lapata-2019-text, cachola-etal-2020-tldr}, opinion summarization cannot rely on reference summaries for model training since it is difficult and expensive to annotate large scale reviews-summary pairs. Also, customers usually care about specific aspects of a product instead of a \textit{general} high-level summary. Thus, fine-grained \textit{aspect-specific} opinion summaries are required, and this makes the annotation process even more difficult and expensive.

\citet{amplayo-etal-2021-aspect} propose an abstractive approach to generate aspect-specific opinion summaries by training on synthetic datasets. They construct synthetic datasets with review elements (words, phrases, or sentences) identified by a multiple instance learning (MIL) module \cite{keeler1991self} learned with silver-standard labels obtained using aspect seed words. We first follow this direction to propose a more straightforward and effective method that excludes the complex learning module to identify aspect-related elements to construct synthetic datasets. Moreover, aspect seed words, which again require human efforts, may not always be available when moving to new domains. Thus we propose another more general solution without the curation and supervision of aspect seed words.



Specifically, we propose two simple yet effective methods to identify aspect-related review sentences and construct aspect-specific synthetic datasets in a \textit{\underline{L}eave-\underline{O}ne-\underline{O}ut} (\textsc{LOO}) \cite{brazinskas-etal-2020-unsupervised, elsahar-etal-2021-self, brazinskas-etal-2022-efficient} style and then finetune pretrained language models (PLMs) on the synthetic datasets: (a) \textsc{SW-LOO} identifies aspect-related sentences by simply exact-matching aspect seed words and outperforms existing approaches by $3.4$ ROUGE-L points and $0.5$ ROUGE-1 point on aspect opinion summaries of \textsc{Space} and \textsc{Oposum+} respectively; (b) \textsc{NLI-LOO} identifies aspect-related sentences with a finetuned NLI \cite{bowman-etal-2015-large, williams-etal-2018-broad} model. Being the first approach that does not use aspect seed words, it outperforms existing approaches on aspect opinion summarization by $1.2$ ROUGE-L points for \textsc{Space} and falls behind at most $1$ ROUGE point on other metrics.



\section{Problem Formulation}

Let $C$ denote a corpus of reviews on entities $\{ e_1, e_2, \dots \}$ (products or business). Let $A_e = \{ a_1, a_2, \dots, a_M \}$ denotes a set of aspects (e.g., \textit{food} or \textit{location} for a hotel) that are relevant for the domain of entities. For each entity $e$, we define its review set as $R_e = \{ r_1, r_2, \dots, r_N \}$. Each review $r$ is a collection of sentences $ \{x_1, x_2, \dots\} $ and each sentence $x$ is a sequence of tokens $\{ w_1, w_2, \dots \}$. Each aspect $a$ is represented by a small set of \textit{seed words} (e.g., \textit{meal} or \textit{buffet} for \textit{food} aspect) $ S_a = \{ v_1, v_2, \dots \}$. Our approaches generate two types of opinion summaries: (a) \textit{general} summary that contains salient opinions over \textit{all} aspects of the entities; and (b) \textit{aspect} summary that focuses on only one specific aspect $a \in A_e$.

\section{Synthetic Dataset Construction}

\smallskip
\noindent
\textbf{Leave-One-Out (\textsc{LOO})} \hspace{4pt} 
We construct synthetic datasets in a \textsc{LOO} style: from a pool of review elements (reviews or review sentences), an element is randomly sampled as a \textit{pseudo-summary}, then we select input reviews from the remaining review elements.

\subsection{Seed Words Based \textsc{LOO}}
\label{sec:sw_loo}
\begin{figure}[t!]
\centering
\hspace*{-0.23cm} 
\includegraphics[width=1.025\linewidth]{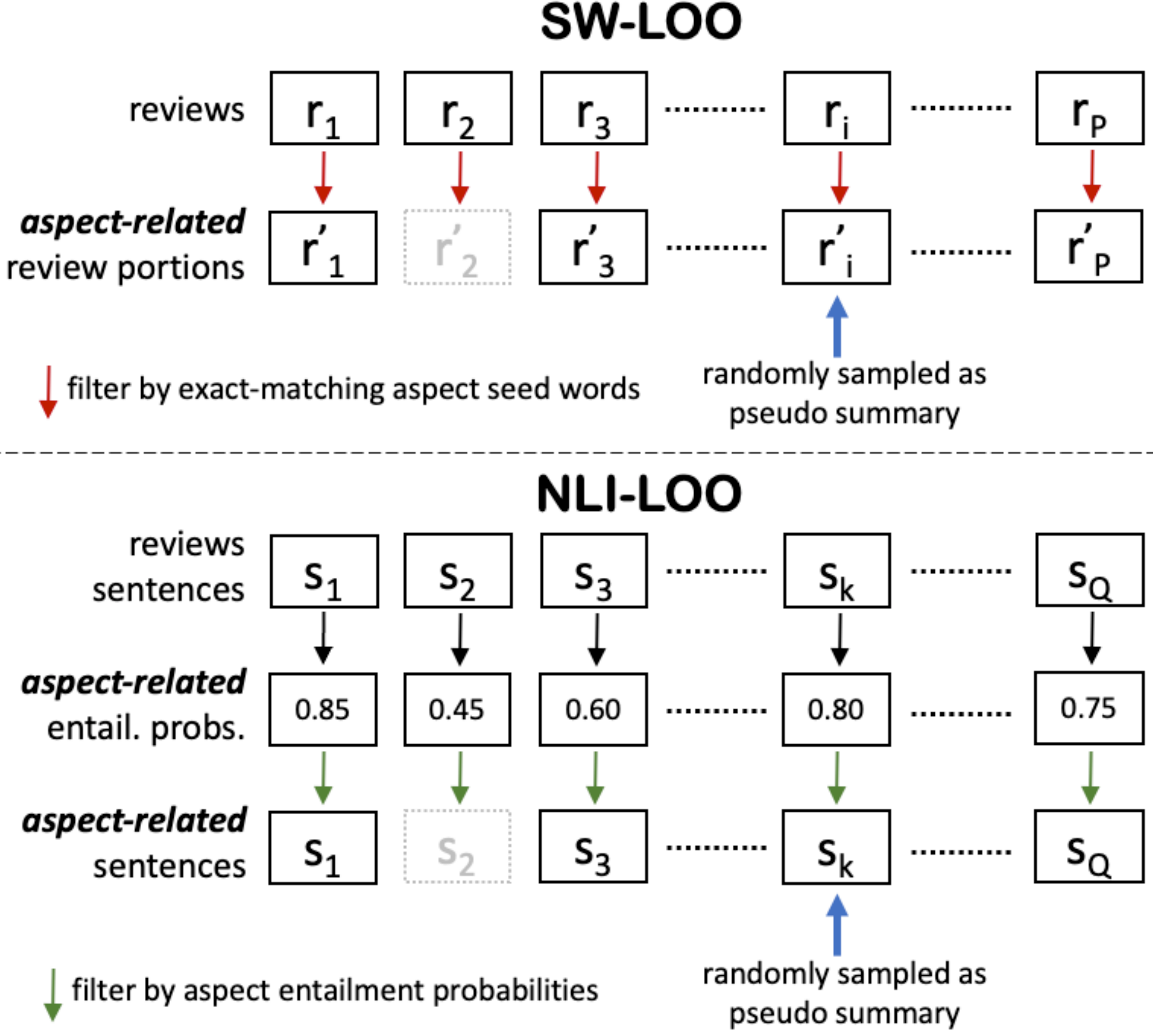}
\caption{\small One synthetic data pair construction for aspect $a$ in \textsc{SW-LOO} and \textsc{NLI-LOO}.} 
\label{fig:methods_figure}
\end{figure}

To build a synthetic reviews-summary pair for aspect $a$, as shown in the upper diagram of Figure \ref{fig:methods_figure}, we first filter each review $r$ into its aspect-related portion $r^{\prime}$ where $r^{\prime} \subseteq \{x_1, x_2, \dots\}$ with each sentence in $r^{\prime}$ containing at least one seed word in $A_e$. For example, for \textit{food} aspect with its seed words \{\textit{breakfast}, \textit{buffet}, ...\}, a hotel review $r_i^{\prime}$: \textit{"They have the most wonderful buffet in Bay Area. And the hotel is close to the airport. Forgot to mention, especially the breakfast is terrific."} will be filtered into its aspect-related review portion $r_i^{\prime}$: \textit{"They have the most wonderful buffet in Bay Area. Forgot to mention, especially the breakfast is terrific."}. Noticed that $r_2^{\prime}$ is empty suppose there is no sentence in $r_2$ containing any seed word. 
Then we apply \textsc{LOO} construction on the filtered aspect-related review portions $\{ r_1^{\prime}, r_3^{\prime}, \dots, r_P^{\prime}\}$ as shown in the diagram: $r_i^{\prime}$ is randomly sampled as the pseudo summary and inputs are chosen from $\{ r_1^{\prime}, r_3^{\prime}, \dots, r_P^{\prime}\} \setminus \{r_i^{\prime}\}$ by first ranking them with the pseudo-summary $r_i^{\prime}$ based on ROUGE-1 score \cite{lin-2004-rouge} and then truncating with a token budget $j$ (truncate up to $j$ tokens) since a concatenation of all filtered reviews cannot fit into the encoder of a PLM. 
Please refer to Appendix \ref{sec:sw_loo_details} for more details and analysis on \textsc{SW-LOO}.

\subsection{NLI Based \textsc{LOO}}
\textbf{NLI Component} \hspace{4pt} In order to relax the requirement of aspect seeds (provided by humans) and to make a more scalable and general solution, we propose to use an NLI model to infer whether a review sentence is related to an aspect. Specifically, we set a review sentence as the premise and verbalize an aspect with the template: \textit{the text is about \{aspect\}}, which we use as the hypothesis. If the entailment probability is higher than a threshold ($0.9$ for \textsc{Space} and $0.8$ for \textsc{Oposum+}), we identify the sentence as related to the aspect with this entailment probability, else we set the aspect-related probability to $0$.

To build a synthetic pair for aspect $a$, we first break all reviews into review sentences and filter out those that are not related to aspect $a$ with the NLI model. As shown in the lower diagram of Figure \ref{fig:methods_figure}, each sentence is first passed through the NLI model to infer its probability of relatedness to aspect $a$, so $s_2$ with entailment probability of $0.45$ will be filtered out if the threshold is set to $0.5$. 
Then we apply \textsc{LOO} construction on all aspect-related sentences $\{ s_1, s_3, \dots, s_Q\}$ and we also use a token budget to truncate ranked synthetic input similar to \textsc{SW-LOO}, however, different from \textsc{SW-LOO} where we use ROUGE-1 scores to rank, we calculate similarities based on entailment probabilities. Please refer to Appendix \ref{sec:nli_loo_details} for more details and analysis on \textsc{NLI-LOO}. Note that we filter the input reviews at sentence level for \textsc{NLI-LOO} and at review level in \textsc{SW-LOO}.

\section{Summarization Model} 
We use \textsc{T5} \cite{raffel2020exploring}, a sequence-to-sequence Transformer-based \cite{NIPS2017_3f5ee243} PLM, to finetune our synthetic datasets similar to previous works \cite{ke2022consistsum, amplayo-etal-2021-aspect}. For \textsc{SW-LOO}, we use the following template: \texttt{``summarize based on aspect: [ASPECT]} \textit{\{aspect\}} \texttt{[ASPECT] with seed words: [SEED]} \textit{\{seed words\}} \texttt{[SEED]:} \textit{\{filtered review\}} \texttt{[SEP]} \textit{\{filtered review\}} $\dots$ \texttt{''} to convert synthetic input and for \textsc{NLI-LOO}, we use: \texttt{``[ASPECT]} \textit{\{aspect\}} \texttt{[SEP]} \textit{\{aspect-related sent\}} \texttt{[SEP]} \textit{\{aspect-related sent\}} $\dots$ \texttt{''}. \texttt{[ASPECT]}, \texttt{[SEED]}, and \texttt{[SEP]} are special tokens, \textit{\{aspect\}} is an aspect name, \textit{\{seed words\}} are concatenation of seed words for an aspect, each \textit{\{filtered review\}} is a $r_i^{\prime}$ in \textsc{SW-LOO} synthetic input, and each \textit{\{aspect-related sent\}} is a $s_k$ in \textsc{NLI-LOO} synthetic input. For both methods, outputs are pseudo summaries.

\section{Experiment}

\subsection{Datasets}
We evaluate our methods on two opinion summarization datasets: \textsc{Space} \cite{angelidis-etal-2021-extractive}, containing reviews from \textit{hotel} domain, and \textsc{Oposum+} \cite{amplayo-etal-2021-aspect}, containing Amazon product reviews from six different domains. Both datasets are comprised of a large corpus of raw reviews and a small development and test set with human-annotated aspects and general opinion summaries for evaluation. Aspect seed words are usually obtained with a small amount of human effort. For \textsc{SW-LOO}, we use the same seed words as in \citet{amplayo-etal-2021-aspect} (Appendix \ref{sec:seed_words_lists}). Refer to Appendix \ref{sec:dataset_details} for detailed descriptions and statistics of the two datasets.

\begin{table}[t]
\hspace*{-0.3cm} 
\centering
\scalebox{0.75}{
    \begin{tabular}{@{}clcccccc@{}}
    \toprule
    \multicolumn{2}{c}{} & \multicolumn{3}{c}{\textsc{Space}} & \multicolumn{3}{c}{\textsc{Oposum+}} \\
    \multicolumn{2}{c}{Model} & R1 & R2 & RL & R1 & R2 & RL \\ 
     \midrule

     \parbox[t]{0.05em}{\multirow{6}{*}{\rotatebox[origin=c]{90}{\small Extractive}}} & 
    \textsc{LexRank} & 24.61 & 3.41 & 18.03 & 22.51 & 3.35 & 17.27 \\
     & \textsc{QT} & 28.95 & 8.34 & 21.77 & 23.99 & 4.36 & 16.61 \\ 
     & \textsc{AceSum\textsubscript{Ext}} & 30.91 & 8.77 & 23.61 & 26.16 & 5.75 & 18.55 \\
     & \textsc{SemAE} & 31.24 & 10.43 & 24.14 & - & - & - \\
     & \textbf{\textsc{SW-LOO\textsubscript{Ext}}} & \underline{33.14} & 10.32 & 25.81 & 28.14 & 6.10 & 19.51 \\
     & \textbf{\textsc{NLI-LOO\textsubscript{Ext}}} & 27.18 & 6.63 & 20.60 & 26.78 & 6.48 & 18.07 \\
    
    \midrule

     \parbox[t]{0.05em}{\multirow{5}{*}{\rotatebox[origin=c]{90}{\small Abstractive}}} & 
     \textsc{MeanSum} & 25.68 & 4.61 & 18.44 & 24.63 & 3.47 & 17.53 \\
     & \textsc{CopyCat} & 27.19 & 5.63 & 19.18 & 26.17 & 4.30 & 18.20 \\
     & \textsc{AceSum} & 32.41 & 9.47 & 25.46 & \underline{29.53} & \underline{6.79} & \textbf{21.06} \\
     & \textbf{\textsc{SW-LOO}} & \textbf{34.68} & \textbf{11.50} & \textbf{28.83} & \textbf{30.00} & \textbf{6.92} & \underline{20.76} \\
     & \textbf{\textsc{NLI-LOO}} & 31.57 & \underline{10.44} & \underline{26.66} & 28.90 & 6.60 & 20.11 \\

    \midrule
     & \textsc{Human} & 44.86 & 18.45 & 34.58 & 43.03 & 16.16 & 31.53 \\
    \bottomrule
    \end{tabular}
}
\caption{\small Evaluation for \textit{aspect summaries} on \textsc{Space} and \textsc{Oposum+} test sets. Best performances are  in \textbf{bold} and second best performances are \underline{underlined}.}
\label{tab:aspect_summaries}
\end{table}

\subsection{Baselines}
We compare our methods with several unsupervised extractive and abstractive approaches. Extractive approaches include \textsc{Centroid} \cite{Radev2004CentroidbasedSO}, \textsc{LexRank} \cite{Erkan2004LexRankGL}, \textsc{QT} \cite{angelidis-etal-2021-extractive}, \textsc{SemAE} \cite{basu-roy-chowdhury-etal-2022-unsupervised}, and two extractive variants of our methods, \textsc{SW-LOO\textsubscript{Ext}} and \textsc{NLI-LOO\textsubscript{Ext}}, by feeding identified aspect-related sentences to \textsc{LexRank} instead of \textsc{T5}, similar to the idea in \citet{amplayo-etal-2021-aspect}. Abstractive approaches include \textsc{MeanSum} \cite{chu2019meansum}, \textsc{CopyCat} \cite{brazinskas-etal-2020-unsupervised}, and \textsc{AceSum} \cite{amplayo-etal-2021-aspect}. 
Appendix \ref{sec:baseline_details} contains more details on baselines.

We also compare with two upper bounds reported in \citet{amplayo-etal-2021-aspect}: an \textsc{Oracle} that selects the review with the highest ROUGE score to the gold summary as the summary and a \textsc{Human} upper bound that is calculated as the inter-annotator ROUGE scores.

\subsection{Implementation} 
\label{sec:implementation}
We first pre-process the raw corpus such as removing products with very few reviews and too long or short reviews as in Appendix \ref{sec:datasets_preprocess}. We use \textsc{T5-Small} as our summarization models and larger \textsc{T5} size does not show improvements as shown in Appendix \ref{sec:t5_size}. We use a \textsc{MNLI} \cite{williams-etal-2018-broad} finetuned \textsc{BART-Large} \cite{lewis-etal-2020-bart} model in \textsc{NLI-LOO}. We choose this model given its better performance\footnote{\url{https://joeddav.github.io/blog/2020/05/29/ZSL.html}} in zero-shot topic classification. We perform simple hyper-parameter tuning on dev sets and select checkpoints with the best ROUGE-L scores to report performances on test sets. Please refer to Appendix \ref{sec:implementation_detials} for more details such as training configurations and other analyses.

\subsection{Results}
We evaluate the quality of generated opinion summaries using ROUGE1/2/L F1 scores \cite{lin-2004-rouge}. Example summaries generated by our methods are shown in Table \ref{tab:space_summaries} and Table \ref{tab:oposum_summaries} in Appendix.

\begin{table}[t]
\hspace*{-0.3cm}
\centering
\scalebox{0.75}{
    \begin{tabular}{@{}clcccccc@{}}
    \toprule
    \multicolumn{2}{c}{} & \multicolumn{3}{c}{\textsc{Space}} & \multicolumn{3}{c}{\textsc{Oposum+}} \\
    \multicolumn{2}{c}{Model} & R1 & R2 & RL & R1 & R2 & RL \\ 
    \midrule
    \parbox[t]{0.05em}{\multirow{7}{*}{\rotatebox[origin=c]{90}{\small Extractive}}} & 
     \textsc{Centroid} & 31.29 & 4.91 & 16.43 & 33.44 & 11.00 & 20.54 \\
     & \textsc{LexRank} & 31.41 & 5.05 & 18.12 & 35.42 & 10.22 & 20.92 \\
     & \textsc{QT} & 38.66 & 10.22 & 21.90 & 37.72 & 14.65 & 21.69 \\ 
     & \textsc{AceSum\textsubscript{Ext}} & 35.50 & 7.82 & 20.09 & 38.48 & 15.17 & 22.82 \\
     & \textsc{SemAE} & \textbf{43.46} & \textbf{13.48} & \textbf{26.40} & - & - & - \\
     & \textbf{\textsc{SW-LOO\textsubscript{Ext}}} & 38.44 & 11.01 & \underline{25.62} & \textbf{40.45} & \textbf{19.13} & \underline{23.20} \\
     & \textbf{\textsc{NLI-LOO\textsubscript{Ext}}} & 25.07 & 4.52 & 16.16 & \underline{39.79} & \underline{18.33} & \textbf{23.49}  \\
    \midrule
    \parbox[t]{0.05em}{\multirow{5}{*}{\rotatebox[origin=c]{90}{\small Abstractive}}} &
     \textsc{MeanSum} & 34.95 & 7.49 & 19.92 & 26.25 & 4.62 & 16.49 \\
     & \textsc{CopyCat} & 36.66 & 8.87 & 20.90 & 27.98 & 5.79 & 17.07 \\
     & \textsc{AceSum} & 40.37 & 11.51 & 23.23 & 32.98 & 10.72 & 20.27 \\
     & \textbf{\textsc{SW-LOO}} & \underline{42.27} & \underline{12.99} & 23.47 & 36.19 & 12.17 & 21.11 \\
     & \textbf{\textsc{NLI-LOO}} & 41.25 & 12.79 & 24.31 & 31.22 & 9.93 & 19.08 \\
    \midrule
     & \textsc{Oracle} & 40.23 & 13.96 & 23.46 & 41.88 & 21.52 & 29.30 \\
     & \textsc{Human} & 49.80 & 18.80 & 29.19 & 55.42 & 37.26 & 44.85 \\
    \bottomrule
    \end{tabular}
}
\caption{\small Evaluation for \textit{general summaries} on \textsc{Space} and \textsc{Oposum+} test sets. Best performances are highlighted in bold and second-best performances are underlined.}
\label{tab:general_summaries}
\end{table}

\smallskip
\noindent
\paragraph{Aspect Opinion Summarization} Table \ref{tab:aspect_summaries} contains the results of all baselines and our methods on the two benchmark datasets. Despite its simplicity, \textsc{SW-LOO} achieves the highest scores on both datasets across all metrics except RL for \textsc{Oposum+} with only $0.3$ points behind the best-performing baseline. On the other hand, \textsc{NLI-LOO} achieves higher R2 and RL scores on \textsc{Space} than existing methods despite using no seed words. While it falls behind other methods on \textsc{Oposum+}, it is at most $1$ point behind across all metrics. This highlights that even without aspect seed words, \textsc{NLI-LOO} is possible to compete with SOTA aspect-based opinion summarization methods.

Next, we turn to the evaluation of extractive versions of our methods. We observe \textsc{SW-LOO\textsubscript{Ext}} achieves higher R1 and RL scores on \textsc{Space} but falls behind on \textsc{Oposum+} by at most $1.5$ point compared with all baselines. This is consistent with the finding in \citet{amplayo-etal-2021-aspect} that a simple centrality-based extractive approach such as \textsc{LexRank} are strong baselines as long as input sentences are already aspect-related. And \textsc{SW-LOO\textsubscript{Ext}} outperforming \textsc{AceSum\textsubscript{Ext}} further shows that our simple filtering method using exact-matching seed words already produces good enough aspect-related sentences compared with the extra learning module used in \citet{amplayo-etal-2021-aspect}. However, \textsc{NLI-LOO\textsubscript{Ext}}, is not able to outperform the best baseline, and we hypothesize the reason is that NLI model filtered aspect-related sentences are still noisy so that a summarization model is required to serve as regularization.

Finally, comparing our four methods, \textsc{SW-LOO} achieves the best performances with the supervision of seed words, \textsc{NLI-LOO} comes second despite the lack of seed words supervision, and our two extractive versions come last since the ground truth summaries are in nature abstractive.

\smallskip
\noindent
\textbf{General Opinion Summarization} \hspace{4pt} 
As shown in Table \ref{tab:general_summaries}, on \textsc{Space}, \textsc{SW-LOO} and \textsc{NLI-LOO} outperform the SOTA abstractive system, \textsc{AceSum}, but under-perform SOTA extractive system, \textsc{SemAE}. We observe the same trend between \textsc{SW-LOO\textsubscript{Ext}} and \textsc{AceSum\textsubscript{Ext}} as in aspect opinion summarization and this again shows the simple yet effective nature of our filtering method. For \textsc{Oposum+}, \textsc{SW-LOO\textsubscript{Ext}} and \textsc{NLI-LOO\textsubscript{Ext}} outperform existing methods given that the annotated general summaries for \textsc{Oposum+} are extractive, \textsc{SW-LOO} outperforms existing abstractive approaches, and \textsc{NLI-LOO} falls behind with only $1$ point.

\begin{table}[t]
\hspace*{-0.2cm}
\centering
\scalebox{0.75}{
    \begin{tabular}{@{}lcccc@{}}
    \toprule
     & \multicolumn{2}{c}{\textsc{Space}} & \multicolumn{2}{c}{\textsc{Oposum+}} \\
    \multicolumn{1}{c}{Model} & Aspect & General & Aspect & General \\ 
    \midrule
    \textsc{SW-LOO} & 27.59 & 23.42 & 20.41 & 20.58 \\
    \multicolumn{1}{l}{\hspace{2pt} \textit{w/ Training Random}} & 24.24 & 24.70 & 19.75 & 18.71 \\
    \multicolumn{1}{l}{\hspace{2pt} \textit{w/ Inference Random}} & 23.46 & 22.04 & 18.76 & 19.41 \\
    \multicolumn{1}{l}{\hspace{2pt} \textit{w/ Both Random}} & 14.71 & 21.82 & 18.06 & 18.15 \\
    \midrule
    \textsc{NLI-LOO} & 25.92 & 25.13 & 19.21 & 19.32 \\
    \multicolumn{1}{l}{\hspace{2pt} \textit{w/ Training Random}} & 22.05 & 22.06 & 18.42 & 19.37 \\
    \multicolumn{1}{l}{\hspace{2pt} \textit{w/ Inference Random}} & 24.33 & 24.56 & 18.10 & 16.97 \\
    \multicolumn{1}{l}{\hspace{2pt} \textit{w/ Both Random}} & 16.14 & 22.50 & 17.83 & 19.69 \\
    \bottomrule
    \end{tabular}
}
\caption{\small \textit{Training Random} means randomly selecting sentences as pseudo summary and input during synthetic dataset construction. \textit{Inference Random} means randomly selecting sentences as input during inference. We report RL scores of our approaches on dev sets.}
\label{tab:ablation}
\end{table}

\subsection{Ablation Study}
We conduct ablation experiments with random filtering to study the importance of the filtering strategies in our two methods. We introduce randomness in two different phases. First, when constructing synthetic pairs, instead of using our filtering strategies before applying \textsc{LOO} construction, we randomly select sentences as pseudo-summary and input. This is essentially a random \textsc{LOO} baseline. Second, during inference, we sample random sentences to feed into \textsc{T5} encoder instead of using our filtering strategies to select aspect-related elements. Finally, we combine these two random strategies. Results in Table \ref{tab:ablation} show that our sentence filtering strategies are crucial since ROUGE scores drop drastically as more randomness is introduced. This is more severe for aspect summarization since aspect-specific synthetic dataset construction needs to focus on particular aspects. However, randomly selecting sentences is possible to cover most aspects by chance for general summarization.

\section{Conclusion}
In this work, we propose two simple yet effective unsupervised approaches that generate aspect and general opinion summaries by training on synthetic datasets. \textsc{SW-LOO} constructs synthetic datasets simply by exact-matching aspect seed words and outperforms existing methods consistently on all metrics and datasets. Being the first work that generates aspect summaries without using aspect seed words, \textsc{NLI-LOO} constructs synthetic datasets with an out-of-the-box NLI model and achieves on-par and sometimes even better performances compared with existing methods.

\section*{Limitations}
One of the biggest challenge in opinion summarization is the multi-document setting where each document represents one product review. Since the number of reviews for a product tends to be large, it would be unrealistic to concatenate all input reviews and train to generate a summary in an end-to-end fashion limited by modern hardware capacity, for example, the GPU memory needed is quadratic w.r.t the input length for all transformer-based PLM. In this work, we tackle this problem by pre-filtering reviews using some heuristics (aspect seed words matching and NLI model selecting) into sub-elements of reviews with much smaller sizes. However, information is very likely to get lost and become incomplete in the pre-filtering phase, leading to inaccurate summarization. Our approach is exactly facing this problem. One way to address this drawback is to first condense each review into an encoding that contains key information of the review such as opinion aspect and opinion sentiment, and then aggregate all review vectors to generate a summary. \citet{amplayo-lapata-2021-informative} call this pipeline as \textsc{Condense-Abstract} and it has been used in both supervised and unsupervised general opinion summarization \cite{chu2019meansum, coavoux-etal-2019-unsupervised, iso-etal-2021-convex-aggregation, amplayo-lapata-2021-informative, isonuma-etal-2021-unsupervised}.


\bibliography{custom}
\bibliographystyle{acl_natbib}

\clearpage

\appendix

\section{Related Works}
Unsupervised opinion summarization is the task of summarizing opinionated text such as customer reviews without training on gold reviews-summary pairs. Recent works have been using autoencoders \cite{Kingma2014AutoEncodingVB} and synthetic datasets construction, or a mix of both, to tackle the zero-shot setting. 

An autoencoder model consists of an encoder that maps the input into latent embedding space and a decoder that reconstructs the original input from the latent space. The latent representation learned can be later aggregated or can be used to cluster and select text to perform both extractive and abstractive summarization. \citet{chu2019meansum, brazinskas-etal-2020-unsupervised} aggregate the input reviews latent representations by averaging then generate the summaries conditioned on it. \citet{angelidis-lapata-2018-summarizing} utilizes the latent representation with aspect specificity and sentiment polarity to guide the selection of review texts as extractive summaries. Recently, \citet{angelidis-etal-2021-extractive} proposes the first approach that generates both general and \textit{aspect-specific} opinion summaries in an extractive manner. They first leverage Vector-Quantized Variational Autoencoder \cite{van2017neural} to cluster review sentences and then use a popularity-driven extraction algorithm to summarize. Similar to \citet{angelidis-etal-2021-extractive}, \citet{basu-roy-chowdhury-etal-2022-unsupervised} learns representations of texts over latent semantic units using dictionary learning \cite{dumitrescu2018dictionary}. Other autoencoder-related methods include denoising autoencoder \cite{amplayo-lapata-2020-unsupervised} and \citet{coavoux-etal-2019-unsupervised}, an encoder-decoder architecture that utilizes clustering of encoding space to extract summaries.

Another direction of work creates synthetic datasets utilizing the largely available amount of online customer reviews. Synthetic datasets are usually constructed in a \textit{leave-one-out} (\textsc{LOO}) style that one review is first randomly sampled as a pseudo-summary, and then a subset of reviews are selected or generated as input reviews to be paired with the pseudo-summary to enable supervised training. Methods of selecting and generating input reviews include random sampling \cite{brazinskas-etal-2020-unsupervised}, generating noisy versions of the pseudo-summary \cite{amplayo-lapata-2020-unsupervised}, selecting reviews that have closer distribution with the pseudo-summary in the embedding space \cite{amplayo2021unsupervised, ke2022consistsum}, and selecting more textual similar reviews \cite{elsahar-etal-2021-self, brazinskas-etal-2022-efficient}. Recently, \citet{amplayo-etal-2021-aspect} proposes the first abstractive approach that can generate both general and aspect summaries. Their method build synthetic datasets by identifying aspect-specific elements with a multiple instance learning (MIL) model \cite{keeler1991self} using aspect seed words. Our work is closest to \citet{amplayo-etal-2021-aspect} in that we also build synthetic datasets by identifying aspect-specific elements, however, our methods do not require extra learning components such as MIL but achieve better performances.

Besides unsupervised opinion summarization, our second method, \textsc{NLI-LOO} is related to the recent approach \cite{yin-etal-2019-benchmarking} that utilizes NLI \cite{bowman-etal-2015-large, williams-etal-2018-broad} models to tackle zero-shot text classification \cite{Chang2008ImportanceOS} (multi-class and multi-label) problem such as topic detection \cite{NIPS2015_250cf8b5} and emotion detection \cite{bostan-klinger-2018-analysis}. The main idea is to solve the classification problem by casting the problem into NLI format. Specifically, the text to be classified becomes the premise, and class labels are converted into natural language format (verbalization) to be used as the hypothesis. If the text entails the verbalized class label, then the text belongs to this class. In our work, we identify the relatedness of a review sentence to an aspect in such a way to construct synthetic datasets.

\section{\textsc{SW-LOO} Details}
\label{sec:sw_loo_details}
For general synthetic pairs construction, after filtering each review with seed words for each aspect, we make sure to sample one review such that its aspect-related portions for all aspects are non-empty and concatenate them as pseudo-summary. We retrieve top similar filtered reviews to each aspect-related portion in the pseudo-summary and concatenate them as general synthetic input, and the retrieval process is the same as in aspect synthetic pairs construction. General synthetic input and output are both approximately $M$ times the length of those in aspect synthetic pairs where $M$ is the number of aspects. For the summarization model, \textit{\{aspect\}} and \textit{\{seed words\}} are the concatenation of all aspects and all seed words for general synthetic pairs. Finally, we train all synthetic pairs together.

At inference time, we also first filter each review into aspect-related portions. However, since there is no reference pseudo summary, we cannot truncate based on similarities to fit into \textsc{T5} encoder. We adopt the \textit{principle strategy} used in PEGASUS \cite{zhang2020pegasus} Gap Sentences Generation pretraining objective to select important reviews as input for inference. We show the effectiveness of adopting the principle strategy in Appendix \ref{sec:principle_strategy}.

\section{\textsc{NLI-LOO} Details}
\label{sec:nli_loo_details}
Different from \textsc{SW-LOO} where we use ROUGE-1 scores, we calculate similarities based on aspect entailment probabilities to rank and truncate aspect-related sentences as synthetic input. For aspect synthetic pairs, we simply calculate the absolute probability difference between pseudo summary and aspect-related sentences. For general synthetic pairs, each review sentence (no matter whether aspect-related) corresponds to a probability vector of dimension $M$ where $M$ is the number of aspects and each element is the probability of the sentence being related to each aspect, and we calculate cosine similarities between the probability vectors of pseudo summary and review sentences that are related to at least one aspect (sum of the probability vector is non-zero). We use the same token budget to truncate review sentences to fit into \textsc{T5} encoder for both aspect and general synthetic pairs. We also train all synthetic pairs together. Another way to calculate similarities is directly using cosine similarity between sentence embeddings, however, results reported in Appendix \ref{sec:nli_similarity} do not show better performance. 

During inference, we use $1$ and all-one vectors with dimension $M$ as reference vectors to rank and truncate review sentences for aspect and general input construction.

\section{Datasets Details}
\label{sec:dataset_details}
\textit{Hotel} reviews in \textsc{Space} are collected from TripAdvisor and each hotel in the evaluation sets is annotated with seven types of summaries: six aspect-specific and one general, with three gold summaries for each type. The number of reviews for a hotel in the raw corpus varies but each hotel in the evaluation sets comes with $100$ reviews. Product reviews from six domains: \textit{laptop bag}, \textit{Bluetooth headset}, \textit{boots}, \textit{keyboard}, \textit{television}, and \textit{vacuum} in \textsc{Oposum+} are initially down-sampled from \textit{Amazon Product Dataset}\footnote{\url{http://jmcauley.ucsd.edu/data/amazon/}} \cite{10.1145/2766462.2767755} by \citet{angelidis-lapata-2018-summarizing} and then further expanded by \citet{amplayo-etal-2021-aspect}. Each product in the evaluation sets is annotated with four types of summaries: three aspect-specific and one general, with also three gold summaries for each type. The number of reviews for a product in the raw corpus also varies but each product in the evaluation sets comes with $10$ reviews. All human-annotated summaries are abstractive except that general summaries in \textsc{Oposum+} are extractive. Detailed statistics of the datasets are shown in Table \ref{tab:datsets_stats}.

\begin{table}[t]
\centering
\hspace*{-0.5cm} 
\scalebox{0.85}{
    \begin{tabular}{@{}lcc@{}}
    \toprule
    Statistics & \textsc{Space} & \textsc{Oposum+} \\
    \midrule
    domain & 1 & 6 \\
    aspects per entity & 6 & 3 \\ 
    \midrule
    \textit{raw review corpus} & \\
    \hspace{8pt} entities & 11.40K & 95.55K \\
    \hspace{8pt} total reviews & 1.14M & 4.13M \\
    \midrule
    \textit{dev / test set} & \\
    \hspace{8pt} entities & 25 & 30 \\ 
    \hspace{8pt} reviews per entity & 100 & 10 \\
    \hspace{8pt} summaries per entity & 3 & 3 \\
    \hspace{8pt} total aspect summaries & 450 & 270 \\
    \hspace{8pt} total general summaries & 75 & \underline{90} \\
    \bottomrule
    \end{tabular}
}
\caption{\small Detailed statistic for \textsc{Space} and \textsc{Oposum+} datasets. Note that only gold general summaries for \textsc{Oposum+}, which is underlined in the table, are extractive.}
\label{tab:datsets_stats}
\end{table}

\begin{table}[t!]
\centering
\hspace*{-0.3cm} 
\scalebox{0.9}{
    \begin{tabular}{@{}lc@{}}
        \toprule[1.25pt]
        Aspect & Hotel \\
        \midrule
        \textsf{building} & lobby pool decor gym area \\
        \textsf{cleanliness} & clean spotless garbage dirty stain \\
        \textsf{food} & breakfast food buffet restaurant meal \\
        \textsf{location} & location walk station distance bus \\
        \textsf{rooms} & room bed bathroom shower spacious \\
        \textsf{service} & staff service friendly helpful desk \\
        \bottomrule[1.25pt]
    \end{tabular}
}
\caption{\small Seed words for \textit{hotel} domain in \textsc{Space} dataset.}
\label{tab:space_seed_words}
\vspace{1em}
\hspace*{-0.3cm} 
\scalebox{0.9}{
    \begin{tabular}{@{}lc@{}}
        \toprule[1.25pt]
        Aspect & Laptop Bag \\
        \midrule
        \textsf{looks} & looks color stylish looked pretty \\
        \textsf{quality} & quality material poor broke durable \\
        \textsf{size} & fit fits size big space \\
        \bottomrule[1.25pt]
        \\
        \toprule[1.25pt]
        Aspect & Boots \\
        \midrule
        \textsf{comfort} & comfortable foot hurt ankle comfy \\
        \textsf{looks} & cute look looked fringe style \\
        \textsf{size} & size half big little bigger \\
        \bottomrule[1.25pt]
        \\
        \toprule[1.25pt]
        Aspect & Bluetooth Headset \\
        \midrule
        \textsf{comfort} & ear fit comfortable fits buds \\
        \textsf{ease of use} & easy button simple setup control \\
        \textsf{sound quality} & sound quality hear noise volume \\
        \bottomrule[1.25pt]
        \\
        \toprule[1.25pt]
        Aspect & Keyboard \\
        \midrule
        \textsf{quality} & working months build stopped quality \\
        \textsf{comfort} & feel comfortable feels mushy shallow \\
        \textsf{layout} & key keys delete backspace size \\
        \bottomrule[1.25pt]
        \\
        \toprule[1.25pt]
        Aspect & TV \\
        \midrule
        \textsf{connectivity} & hdmi computer port usb internet \\
        \textsf{image quality} & picture color colors bright clear \\
        \textsf{sound quality} & sound speakers loud tinny bass \\
        \bottomrule[1.25pt]
        \\
        \toprule[1.25pt]
        Aspect & Vacuum \\
        \midrule
        \textsf{accessory} & filter brush attachments attachment turbo \\
        \textsf{ease of use} & easy push concerns awkward impossible \\
        \textsf{suction} & suction powerful power hair quiet \\
        \bottomrule[1.25pt]
    \end{tabular}
}
\caption{\small Seed words for various domains in \textsc{Oposum+} dataset.}
\label{tab:oposum_seed_words}
\end{table}

\section{List of Seed Words}
\label{sec:seed_words_lists}
Aspect seed words (listed in Table \ref{tab:space_seed_words} and \ref{tab:oposum_seed_words}) are usually automatically extracted using a variant of clarity scoring function \cite{CronenTownsend2002PredictingQP} applied on a small amount of aspect annotation as described in \citet{angelidis-lapata-2018-summarizing}, and they can be further manually improved by domain experts as in \citet{amplayo-etal-2021-aspect}.

\section{Baselines Details}
\label{sec:baseline_details}

\textbf{\textit{Extractive Approaches}} \hspace{4pt} We first compare against two traditional approaches: \textsc{Centroid} selects the review closest to the centroid of all reviews as the summary; \textsc{LexRank} selects the most salient review sentences as summary similar to \textsc{PageRank} \cite{Page1999ThePC}. \textsc{BERT} \cite{devlin-etal-2019-bert} embedding is used to represent sentences in both traditional methods. More recent systems include \textsc{QT} (described in Section \ref{sec:intro}) and \textsc{SemAE}. Inspired by \textsc{QT}, \textsc{SemAE} represents text over latent semantic units using dictionary learning. 

\noindent
\textbf{\textit{Abstractive Approaches}} \hspace{4pt} \textsc{MeanSum} generates summaries by reconstructing the mean of reviews' representations using autoencoder. \textsc{CopyCat} uses a hierarchical variational autoencoder to learn latent codes for the summaries. The most recent approach is \textsc{AceSum}. (described in Section \ref{sec:intro}).

Note that \textsc{LexRank}, \textsc{MeanSum}, and \textsc{CopyCat} do not support aspect-specific summary generation, \citet{amplayo-etal-2021-aspect} adopt a simple sentence-filtering strategy to enable it. Specifically, after training a general opinion summarization model, during inference for aspect summaries, they filter out input review sentences that are not aspect-related using cosine similarities scores between \textsc{BERT} embeddings of review sentences and aspect seed words before feeding into general summarization model.

\section{Datasets Pre-Processing}
\label{sec:datasets_preprocess}
We pre-process differently for our two methods on the same dataset since we want to control the constructed synthetic datasets to have reasonable sizes and resemble properties of test time data such as the number of reviews per product and average review length. We use dev sets to observe such properties. In \textsc{SW-LOO}, we first remove reviews with less than $20$ words and then remove hotels with less than $10$ reviews for \textsc{Space}; we first remove reviews less than $20$ or more than $100$ words then remove products with less than $12$ reviews for \textsc{Oposum+}. In \textsc{NLI-LOO}, we remove reviews with less than $10$ or more than $120$ words for \textsc{Space} and remove reviews with less than $20$ or more than $100$ words for \textsc{Oposum+}.

\begin{table}[t]
\centering
\hspace*{-0.5cm} 
\scalebox{0.85}{
    \begin{tabular}{@{}lrl@{}}
    \toprule[1.25pt]
    \multicolumn{3}{c}{\textsc{SW-LOO}} \\
    \midrule 
    \multirow{2}{*}{\textsc{Space}} & asp. & \texttt{lr=$3\mathrm{e}{-4}$, bch=$16$, bm=$2$} \\
    & gen. & \texttt{lr=$3\mathrm{e}{-4}$, bch=$16$, bm=$2$} \\
    \midrule
    \multirow{2}{*}{\textsc{Oposum+}} & asp. & \texttt{lr=$3\mathrm{e}{-4}$, bch=$16$, bm=$2$} \\
    & gen. & \texttt{lr=$1\mathrm{e}{-6}$, bch=$16$, bm=$2$} \\
    \midrule[1.25pt]
    \multicolumn{3}{c}{\textsc{NLI-LOO}} \\
    \midrule 
    \multirow{2}{*}{\textsc{Space}} & asp. & \texttt{lr=$4\mathrm{e}{-5}$, bch=$16$, bm=$2$} \\
    & gen. & \texttt{lr=$4\mathrm{e}{-5}$, bch=$16$, bm=$2$} \\
    \midrule
    \multirow{2}{*}{\textsc{Oposum+}} & asp. & \texttt{lr=$3\mathrm{e}{-4}$, bch=$8$, bm=$4$} \\
    & gen. & \texttt{lr=$1\mathrm{e}{-6}$, bch=$16$, bm=$2$} \\
    \midrule[1.25pt]
    \multicolumn{3}{c}{\textsc{SW-LOO\textsubscript{Ext}}} \\
    \midrule
    \multirow{2}{*}{\textsc{Space}} & asp. & \texttt{\textsc{BERT-Base}, n=$2$} \\
    & gen. & \texttt{\textsc{BERT-Base}, n=$2$} \\
    \midrule
    \multirow{2}{*}{\textsc{Oposum+}} & asp. & \texttt{\textsc{BERT-Base}, n=$2$} \\
    & gen. & \texttt{\textsc{BERT-Large}, n=$2$} \\
    \midrule[1.25pt]
    \multicolumn{3}{c}{\textsc{NLI-LOO\textsubscript{Ext}}} \\
    \midrule
    \multirow{2}{*}{\textsc{Space}} & asp. & \texttt{\textsc{BERT-Large}, n=$2$} \\
    & gen. & \texttt{\textsc{BERT-Base}, n=$4$} \\
    \midrule
    \multirow{2}{*}{\textsc{Oposum+}} & asp. & \texttt{\textsc{BERT-Large}, n=$4$} \\
    & gen. & \texttt{\textsc{BERT-Large}, n=$4$} \\
    \bottomrule[1.25pt]
    \end{tabular}
}
\caption{\small Best hyper-parameter settings on \textsc{Space} and \textsc{Oposum+} dev sets: \texttt{lr} stands for \texttt{AdamW} initial learning rate, \texttt{bch} stands for training batch size, and \texttt{bm} stands for beam search size at inference time.}
\label{tab:best_hyper_parameter}
\end{table}

\section{Implementation Details}
\label{sec:implementation_detials}
We use \textsc{T5} implementation from HuggingFace\footnote{\url{https://huggingface.co/docs/transformers/model_doc/t5}} \cite{wolf-etal-2020-transformers}. We use \texttt{AdamW} \cite{Loshchilov2019DecoupledWD} optimizer without weight decay and set $0.9$, $0.999$, $1 \times 10\textsuperscript{-8}$ for $\beta\textsubscript{1}$, $\beta\textsubscript{2}$, $\epsilon$. We train all summarization models for a total of $25$K steps on the combination of aspect and general synthetic pairs. We set ngram refraining size \cite{paulus2018a} to $3$ during inference. We tune initial learning rate in [$1\mathrm{e}{-6}$, $4\mathrm{e}{-5}$, $3\mathrm{e}{-4}$] and batch size in [$8$, $16$]. We tune beam search size during inference in [$2$, $4$]. For \textsc{SW-LOO\textsubscript{Ext}} and \textsc{NLI-LOO\textsubscript{Ext}}, we use \texttt{[CLS]} token embedding in the last layer of \textsc{BERT} as the sentence representation. We concatenate top $6$ sentences returned by \textsc{LexRank} as general summary, and tune in [$2$, $4$] for aspect summary. We also use two sizes of \textsc{BERT}: \textsc{BERT-Base} and \textsc{BERT-Large}. All computations are performed on 8-GPU p3.16xlarge Amazon instance. The best hyper-parameter settings for all experiments can be found in Table \ref{tab:best_hyper_parameter}.

During preliminary studies for aspect synthetic pairs construction, we find that for \textsc{Space}, using sampled filtered aspect-related review portion as pseudo-summary rather than the original review that contains the pseudo-summary gives better downstream ROUGE scores, but it is the opposite way with \textsc{Oposum+}. Please refer to Appendix \ref{sec:granularity} for analyses on pseudo-summary granularity. 

\smallskip
\noindent
\textbf{\textsc{SW-LOO}} \hspace{4pt} For \textsc{Space}, we add a linear learning rate warm-up in the first $500$ steps and save checkpoints every $500$ steps. Since there are totally $6$ aspects for \textsc{Space} and very few reviews containing seed words for all $6$ aspects can be sampled as pseudo summaries for general synthetic pairs construction, we relax the constraint of pseudo summaries containing seed words for all $6$ aspects to $4$ aspects. We set $200$ as the token budget to truncate ranked aspect-related review portions for aspect synthetic pairs construction, and $150$ as the token budget in principle strategy when selecting important sentences as input during inference for \textsc{Space}. We set $1536$ and $200$ as the maximum input and output token length of \textsc{T5} for all \textsc{SW-LOO} experiments. Notice that this exceeds $512$, which is the maximum token length that \textsc{T5} is pretrained on, but recent works \cite{zhang2020pegasus, rothe-etal-2021-thorough} have shown that seq2seq PLMs generalize well even when finetuned on longer sequences not observed at pretraining phase. For \textsc{Oposum+}, we add linear learning rate warm-up in the first $250$ steps. We set $300$ as the token budget to truncate for aspect synthetic pairs construction. There are $\sim50$K aspect and $\sim5$K general synthetic pairs for \textsc{Space}, $\sim70$K aspect and $\sim6$K general synthetic pairs for \textsc{Oposum+}.

\smallskip
\noindent
\textbf{\textsc{NLI-LOO}} \hspace{4pt} 
For \textsc{Space}, we add linear learning rate warm-up to first $1000$ steps, and $500$ steps for \textsc{Oposum+}. We set $0.9$ and $0.8$ as the entailment probability threshold for \textsc{Space} and \textsc{Oposum+} based on our preliminary experiments (lower thresholds make identified aspect-related sentences too noisy and further hurt downstream ROUGE scores). For summarization models, we set $500$ as the token budget for both aspect and general synthetic pairs construction for both datasets and set $512$ and $150$ for maximum input and output token length of \textsc{T5}. There are $\sim36$K aspect and $\sim6$K general synthetic pairs for \textsc{Space}, $\sim70$K aspect-specific and $\sim28$K general synthetic pairs for \textsc{Oposum+}.

\begin{table}[t]
\centering
\scalebox{0.75}{
    \begin{tabular}{@{}lcccccc@{}}
    \toprule
     & \multicolumn{3}{c}{Aspect} & \multicolumn{3}{c}{General} \\
    \multicolumn{1}{c}{Model} & R1 & R2 & RL & R1 & R2 & RL \\ 
    \midrule
    \textsc{SW-LOO} & 33.11 & 10.98 & 27.59 & 40.26 & 12.04 & 23.42 \\
    \multicolumn{1}{l}{\hspace{2pt} \textit{w/o Prin. Sel.}} & 30.88 & 9.74 & 25.78 & 35.84 & 10.28 & 21.80 \\
    \bottomrule
    \end{tabular}
}
\caption{\small Randomly or using principle strategy to select aspect-related review portions in order to fit into the encoder of \textsc{T5}. Performances are reported on \textsc{Space} dev set.}
\label{tab:principle_selection}
\end{table}

\section{Principle Strategy Effectiveness}
\label{sec:principle_strategy}
Unlike \textsc{Oposum+}, there are $100$ reviews for each hotel in \textsc{Space} evaluation sets. During inference, we cannot simply concatenate all filtered reviews as input since they cannot fit into \textsc{T5} encoder. We adopt the principle strategy introduced in \textsc{Pegasus} to select the most important filtered reviews and concatenate them as input for inference. In Table \ref{tab:principle_selection}, we show the effectiveness of the principle strategy by comparing it with randomly selecting filtered reviews as input for inference.

\begin{table}[t]
\hspace*{-0.3cm}
\centering
\scalebox{0.75}{
    \begin{tabular}{@{}c|lcccccc@{}}
    \toprule
    \multicolumn{2}{c}{} & \multicolumn{3}{c}{\textsc{Space}} & \multicolumn{3}{c}{\textsc{Oposum+}} \\
    \multicolumn{2}{c}{Model} & R1 & R2 & RL & R1 & R2 & RL \\ 
    \midrule
    \parbox[t]{0.8em}{\multirow{2}{*}{\rotatebox[origin=c]{90}{Asp.}}} & 
    \textsc{NLI-LOO} & 30.20 & 9.84 & 25.92 & 27.48 & 5.64 & 19.21  \\
     & \multicolumn{1}{l}{\hspace{2pt} \textit{w/ Sent. Sim.}} & 29.87 & 9.30 & 25.26 & 27.00 & 6.20 & 19.06 \\
    \midrule
    \parbox[t]{0.8em}{\multirow{2}{*}{\rotatebox[origin=c]{90}{Gen.}}} & 
    \textsc{NLI-LOO} & 41.17 & 12.34 & 25.13 & 31.10 & 10.09 & 19.32 \\
     & \multicolumn{1}{l}{\hspace{2pt} \textit{w/ Sent. Sim.}} & 25.01 & 9.68 & 17.34 & 31.11 & 10.43 & 19.86 \\
    \bottomrule
    \end{tabular}
}
\caption{\small Calculate cosine similarity using aspect entailment probability or sentence embeddings when constructing synthetic datasets in \textsc{NLI-LOO}. Performances are reported on dev sets for both datasets.}
\label{tab:nli_sent_emb}
\end{table}

\section{Similarity Metric}
\label{sec:nli_similarity}
Different from using aspect entailment probability, we can also use sentence embeddings \cite{reimers-gurevych-2019-sentence} to calculate the cosine similarity between pseudo summary and aspect-related review sentences to construct synthetic input. Specifically, we use \texttt{all-mpnet-base-v2}\footnote{\url{https://www.sbert.net/docs/pretrained_models.html}}, which is a sentence embedding model finetuned on a $1$B sentence pairs dataset with a self-supervised contrastive learning objective. Results in Table \ref{tab:nli_sent_emb} show that there is no significant difference except general summarization for \textsc{Space} where using sentence embeddings is much worse than using aspect entailment probability.

\begin{table}[t]
\hspace*{-0.5cm}
\centering
\scalebox{0.75}{
    \begin{tabular}{@{}lcccccc@{}}
    \toprule
     & \multicolumn{3}{c}{Aspect} & \multicolumn{3}{c}{General} \\
    \multicolumn{1}{c}{Granularity} & R1 & R2 & RL & R1 & R2 & RL \\ 
    \midrule
    \textsc{Space} & \\
    \multicolumn{1}{l}{\hspace{2pt} Sentence} & 33.11 & 10.98 & 27.59 & 40.26 & 12.04 & 23.42 \\
    \multicolumn{1}{l}{\hspace{2pt} \textit{Review}} & 25.01 & 6.42 & 18.04 & 39.86 & 11.21 & 23.07 \\
    \midrule
    \textsc{Oposum+} & \\
    \multicolumn{1}{l}{\hspace{2pt} Review} & 29.18 & 6.38 & 20.41 & 36.16 & 11.89 & 20.58 \\
    \multicolumn{1}{l}{\hspace{2pt} \textit{Sentence}} & 22.34 & 5.06 & 17.33 & 20.06 & 6.76 & 13.86 \\
    \bottomrule
    \end{tabular}
}
\caption{\small Pseudo summary granularity study for \textsc{SW-LOO} and \textsc{NLI-LOO}. Performances are reported on dev sets. Note that in our main experiments, we use sentence level pseudo summary for \textsc{Space} and review level for \textsc{Oposum+}}
\label{tab:granularity_study}
\end{table}

\section{Pseudo Summary Granularity}
\label{sec:granularity}
We use different pseudo-summary granularity for two datasets: sentence level for \textsc{Space} and review level for \textsc{Oposum+}. Sentence level directly uses sampled filtered aspect-related review portion (\textsc{SW-LOO}) or sampled aspect-related review sentence (\textsc{NLI-LOO}) as pseudo-summary, and review level uses the original review that contains the sampled pseudo-summary as pseudo-summary. Results in Table \ref{tab:granularity_study} show the importance of design choices for synthetic datasets construction.

\begin{table}[t]
\hspace*{-0.4cm}
\centering
\scalebox{0.75}{
    \begin{tabular}{@{}c|lcccccc@{}}
    \toprule
    \multicolumn{2}{c}{} & \multicolumn{3}{c}{\textsc{Space}} & \multicolumn{3}{c}{\textsc{Oposum+}} \\
    \multicolumn{2}{c}{Model} & R1 & R2 & RL & R1 & R2 & RL \\ 
    \midrule
    \parbox[t]{0.8em}{\multirow{8}{*}{\rotatebox[origin=c]{90}{Aspect Summary}}} & 
    \textsc{\small SW-LOO} &  \\
     & \multicolumn{1}{l}{\hspace{2pt} \textsc{T5-small}} & 33.11 & 10.98 & 27.59 & 29.18 & 6.38 & 20.41 \\
     & \multicolumn{1}{l}{\hspace{2pt} \textsc{T5-base}} & 33.43 & 11.08 & 27.73 & 30.03 & 6.60 & 20.53 \\
     & \multicolumn{1}{l}{\hspace{2pt} \textsc{T5-large}} & 33.70 & 10.77 & 27.60 & 28.98 & 6.15 & 20.32 \\
    \cline{2-8}
     & \textsc{\small NLI-LOO} &  \\
     & \multicolumn{1}{l}{\hspace{2pt} \textsc{T5-small}} & 30.20 & 9.84 & 25.92 & 27.48 & 5.64 & 19.21 \\
     & \multicolumn{1}{l}{\hspace{2pt} \textsc{T5-base}} & 30.24 & 10.04 & 25.95 & 27.58 & 5.28	& 19.29 \\
     & \multicolumn{1}{l}{\hspace{2pt} \textsc{T5-large}} & 30.61 & 9.50 & 25.68 & 27.14 & 5.47 & 19.55 \\
    \midrule
    \parbox[t]{0.8em}{\multirow{8}{*}{\rotatebox[origin=c]{90}{General Summary}}} & 
    \textsc{\small SW-LOO} &  \\
     & \multicolumn{1}{l}{\hspace{2pt} \textsc{T5-small}} & 40.26 & 12.04 & 23.42 & 36.16 & 11.89 & 20.58 \\
     & \multicolumn{1}{l}{\hspace{2pt} \textsc{T5-base}} & 41.31 & 12.47 & 23.12 & 35.53 & 11.65 & 20.33 \\
     & \multicolumn{1}{l}{\hspace{2pt} \textsc{T5-large}} & 39.90 & 10.94 & 22.64 & 32.96 & 10.24 & 19.41 \\
    \cline{2-8}
     & \textsc{\small NLI-LOO} &  \\
     & \multicolumn{1}{l}{\hspace{2pt} \textsc{T5-small}} & 41.17 & 12.34 & 25.13 & 31.10 & 10.09 & 19.32 \\
     & \multicolumn{1}{l}{\hspace{2pt} \textsc{T5-base}} & 37.49 & 11.44 & 22.91 & 26.51 & 6.74	& 17.08 \\
     & \multicolumn{1}{l}{\hspace{2pt} \textsc{T5-large}} & 37.57 & 10.14 & 21.77 & 30.41 & 6.77 & 17.37 \\
    \bottomrule
    \end{tabular}
}
\caption{\small Using different \textsc{T5} sizes as summarization model. Performances are reported on dev sets for both datasets.}
\label{tab:t5_sizes}
\end{table}

\section{\textsc{T5} Model Sizes}
\label{sec:t5_size}
We use different \textsc{T5} sizes including \textsc{T5-Small}, \textsc{T5-Base}, and \textsc{T5-Large} as summarization models. Results in Table \ref{tab:t5_sizes} show that larger summarization models do not necessarily guarantee better downstream ROUGE scores and sometimes even hurt downstream performances. Our hypothesis is that larger models overfit synthetic datasets and thus perform slightly worse on downstream evaluation sets.


\begin{table*}[th!]
\small
\centering
\scalebox{1}{
    \begin{tabularx}{\textwidth}{X}
    \toprule[1.5pt]
    \multicolumn{1}{c}{\textsc{SW-LOO} Summaries} \\
    \midrule
    \textbf{\normalsize Building} \hspace{6pt} The pool area was very nice and the room was clean and comfortable.\\
    \midrule
    \textbf{\normalsize Cleanliness} \hspace{6pt} Our room was very clean and comfortable. \\
    \midrule
    \textbf{\normalsize Food} \hspace{6pt} The breakfast was great and the staff was very helpful and helpful. \\
    \midrule
    \textbf{\normalsize Location} \hspace{6pt} The hotel is located right next to the main road and is a short walk from the beach. \\
    \midrule
    \textbf{\normalsize Rooms} \hspace{6pt} The room was very clean and comfortable. \\
    \midrule
    \textbf{\normalsize Service} \hspace{6pt} The staff was very friendly and helpful. \\
    \midrule
    \textbf{\normalsize General} \hspace{6pt} The pool was very nice and clean. We were able to walk to the beach and Duval st. from the hotel, so we had a nice view of the harbor and the sea! The breakfast was great and we stayed in October and were very pleased with the location - right next to all the restaurants ... The room was small but very small and very comfortable with clean and comfortable beds. \\
    \bottomrule[1.5pt]
    \\
    \toprule[1.5pt]
    \multicolumn{1}{c}{\textsc{NLI-LOO} Summaries} \\
    \midrule
    \textbf{\normalsize Building} \hspace{6pt} The hotel is a beautiful old hotel. \\
    \midrule
    \textbf{\normalsize Cleanliness} \hspace{6pt} The room was clean and the staff was very helpful. \\
    \midrule
    \textbf{\normalsize Food} \hspace{6pt} The breakfast was great and the view from the rooftop was amazing. \\
    \midrule
    \textbf{\normalsize Location} \hspace{6pt} The location is great - just a short walk to the Spanish Steps and the metro station. \\
    \midrule
    \textbf{\normalsize Rooms} \hspace{6pt} The rooms are small by European standards, but very clean and comfortable. \\
    \midrule
    \textbf{\normalsize Service} \hspace{6pt} the service was excellent and the staff was very friendly and helpful. \\
    \midrule
    \textbf{\normalsize General} \hspace{6pt} The hotel is very clean and the staff is friendly and helpful. The room was very small and clean, but the bathroom was a bit small compared to the other rooms in the UK. It is OK to stay here again. I would stay there again if you want to go back to Europe! The location is great - the city is just ten minutes walk from the metro station andn't be disappointed with the price of the rooms. \\
    \bottomrule[1.5pt]
    \end{tabularx}
}
\caption{\small \textit{General} and \textit{aspect-level} summaries for a hotel in \textsc{Space} dataset generated by \textsc{SW-LOO} and \textsc{NLI-LOO}}
\label{tab:space_summaries}
\end{table*}

\begin{table*}[th!]
\small
\centering
\scalebox{1}{
    \begin{tabularx}{\textwidth}{X}
    \toprule[1.5pt]
    \multicolumn{1}{c}{\textsc{SW-LOO} Summaries} \\
    \midrule
    \textbf{\normalsize Sound Quality} \hspace{6pt} I love this headset. It's a great product, but it doesn't have any issues with the sound! It is OK if you are looking for something that can be used for your Samsung TV? \\
    \midrule
    \textbf{\normalsize Comfort} \hspace{6pt} I love this headset. It's a great headset for the price, but it doesn't fit my ear perfectly! \\
    \midrule
    \textbf{\normalsize Ease of Use} \hspace{6pt} I bought this for my Motorola. It is very easy to set up, and the buttons are very comfortable! \\
    \midrule
    \textbf{\normalsize General} \hspace{6pt} I haven't found any way of getting that to be consistently good. The earpieces are not as sturdy or high quality in material as a Motorola, but the buttons are quite accessible and the sound varies based on how it's fitting into my ears! The set is very comfortable and has great range ( roughly 100 feet ) and connects easily to my iPhone with me - but it is not too big for me to wear if it doesn’t fit my TV. \\
    \bottomrule[1.5pt]
    \\
    \toprule[1.5pt]
    \multicolumn{1}{c}{\textsc{NLI-LOO} Summaries} \\
    \midrule
    \textbf{\normalsize Sound Quality} \hspace{6pt} I love these headphones. They are very comfortable, sound quality is good and they're very good quality for the price! \\
    \midrule
    \textbf{\normalsize Comfort} \hspace{6pt} I love these headphones. They are very comfortable, and the sound quality is great! They're a little tight on my ears but if you aren't sure how long they will last you... \\
    \midrule
    \textbf{\normalsize Ease of Use} \hspace{6pt} I bought these for my Motoactv. They are very comfortable to wear, and they don't touch my neck at all! \\
    \midrule
    \textbf{\normalsize General} \hspace{6pt} I bought these headphones in a package for the Motoactv. They are very comfortable, the neck band doesn't touch my neck at all allowing for free movement! The sound is very good and fits comfortably in my ears... but it takes some time to find the right angel and fit it right in. \\
    \bottomrule[1.5pt]
    \end{tabularx}
}
\caption{\small \textit{General} and \textit{aspect-level} summaries for a product in "Bluetooth Headset" domain of \textsc{Oposum+} dataset generated by \textsc{SW-LOO} and \textsc{NLI-LOO}.}
\label{tab:oposum_summaries}
\end{table*}

\end{document}